# Use of a Multiscale Vision Transformer to predict Nursing Activities Score from Low Resolution Thermal Videos in an Intensive Care Unit

Isaac YL Lee[1], Thanh Nguyen-Duc[2], Ryo Ueno[2], Jesse Smith[2] and Peter Y Chan[2]

[1] *Monash University, Melbourne, Victoria, Australia*
[2] *Eastern Health, Melbourne, Victoria, Australia*

**Abstract**—Excessive caregiver workload in hospital nurses has been implicated in poorer patient care and increased worker burnout. Measurement of this workload in the Intensive Care Unit (ICU) is often done using the Nursing Activities Score (NAS), but this is usually recorded manually and sporadically. Previous work has made use of Ambient Intelligence (AmI) by using computer vision to passively derive caregiver-patient interaction times to monitor staff workload. In this letter, we propose using a Multiscale Vision Transformer (MViT) to passively predict the NAS from low-resolution thermal videos recorded in an ICU. 458 videos were obtained from an ICU in Melbourne, Australia and used to train a MViTv2 model using an indirect prediction and a direct prediction method. The indirect method predicted 1 of 8 potentially identifiable NAS activities from the video before inferring the NAS. The direct method predicted the NAS score immediately from the video. The indirect method yielded an average 5-fold accuracy of 57.21%, an area under the receiver operating characteristic curve (ROC AUC) of 0.865, a F1 score of 0.570 and a mean squared error (MSE) of 28.16. The direct method yielded a MSE of 18.16. We also showed that the MViTv2 outperforms similar models such as R(2+1)D and ResNet50-LSTM under identical settings.

This study shows the feasibility of using a MViTv2 to passively predict the NAS in an ICU and monitor staff workload automatically. Our results above also show an increased accuracy in predicting NAS directly versus predicting NAS indirectly. We hope that our study can provide a direction for future work and further improve the accuracy of passive NAS monitoring.

## I. INTRODUCTION

The Intensive Care Unit (ICU) is a complex environment where critically ill patients require constant monitoring and care provided by many caregivers in different roles [1, 2]. Caregiver workload, particularly in the ICU setting, is increasingly becoming recognized as an issue affecting patient safety, caregiver health and health provider costs [2]. The complexities of caregiver workload are influenced by various factors such as patient acuity, task complexity, staffing numbers and human factors impacting stress and burnout [2-4]. In a systematic review by Pappa et al., at least 1 in 5 Health Care Workers (HCWs) reported symptoms of depression and anxiety, while 38.9% had insomnia [5]. In another systematic review, al Falasi et al. found that acute post-traumatic stress disorder (PTSD) symptoms were observed in 3.4% to 71.5% of HCWs [6].

Given the contribution of workload burden to burnout, one proposed solution is to increase its monitoring to allow for more even distribution. Currently, the main methods used to calculate nursing workload are the Therapeutic Intervention Scoring System-28 (TISS-28), the Nine Equivalents of Manpower Score (NEMS) and the Nursing Activities Score (NAS) [7-10]. When compared with TISS-28 and NEMS in the ICU setting, the NAS achieved better scores and is most accurate at measuring nursing workload in the ICU [11-13]. Yet, like all other scoring systems, the NAS is recorded manually and typically only once every eight hours.

In recent years, there has been increased attention on using Ambient Intelligence (AmI) to improve clinical workflow [14]. AmI refers to technology that is sensitive, responsive, adaptive, transparent, ubiquitous and intelligent [15]. For instance, using cameras to continuously monitor vital signs of patients without any patient contact [16]. In addition, AmI has been used to observe other subjective patient states such as delirium or pain [17, 18]. More recently, AmI was used to monitor physical interaction times between healthcare staff and patients, with the goal of measuring staff workload to reduce overwork [18, 19]. There is potential for AmI to be used to continuously monitor metrics that are typically manually scored, such as the NAS.

In this letter, we propose two Multiscale Vision Transformer (MViT) methods for the automatic calculation of the NAS in the ICU using low resolution infrared video feeds. The first indirect method predicts the NAS activity and infers the NAS from it. The second direct method predicts the NAS immediately without interpreting individual actions. Using 458 of our self-annotated ICU videos, we also compare the MViT to traditional models such as R(2+1)D and ResNet50-LSTM under identical settings.

## II. MATERIALS AND METHODS

Figure 1 provides an overview of our research workflow. From a comprehensive dataset of thermal videos captured in the ICU, 882 videos were chosen at random. Two independent clinicians then manually labelled each video with 1 of 23 possible NAS activities observed in the footage. Following this labelling process, the videos underwent preprocessing to prepare them for model training. In the final step, the trained model was used to predict the NAS.

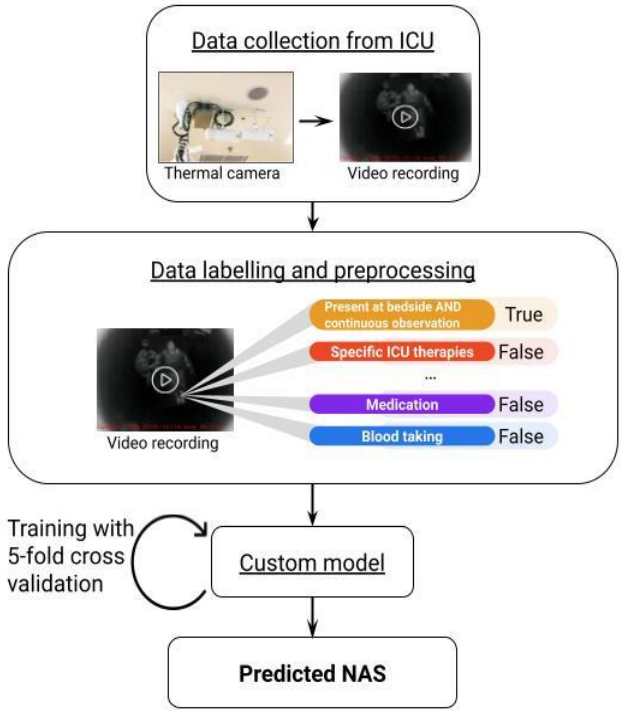

Fig. 1. Research workflow for the prediction of NAS from an ICU video using a custom model (i.e. MViTv2, R(2+1)D or ResNet50-LSTM).

Table 1: The 14 NAS activities and their no. of occurrences (before and after preprocessing).

| NAS activity | No. of occurrences | | Average NAS[b] |
|---|---|---|---|
| | Before | After[a] | |
| Present at bedside AND continuous observation | 88 | 65 | 12.07 |
| Specific ICU therapies | 63 | 58 | 2.80 |
| Adjusting or manipulating NG tube/feeds | 20 | | |
| Checking or insertion of urinary catheter | 11 | | |
| Intratracheal suctioning | 9 | | |
| Care of artificial airways | 34 | | |
| Respiratory Support | 49 | | |
| Processing of clinical data | 85 | 68 | 19.13 |
| Support/interaction with relatives | 57 | 54 | 18.00 |
| Mobilisation and positioning | 73 | 60 | 11.63 |
| Care of drains | 23 | | |
| Hygiene procedure | 54 | 46 | 13.53 |
| Medication | 59 | 57 | 5.60 |
| Blood taking | 55 | 50 | 4.30 |

[a] As only videos with exactly 1 occurrence are included after preprocessing, the number of occurrences is equal to the number of videos.
[b] Average NAS = Average NAS across all subcategories of that activity

### A. Data Collection and Labelling

882 infrared thermal videos were collected as part of an observational study in the ICU ward of Box Hill Hospital, Melbourne from 21 February 2019 to 18 September 2020.

Videos were captured using a Thermal Experts TE-Q1 narrow-angle camera (Daejon, Korea) mounted on the ceiling at a 30° downward angle, 2.2m away from the patient. The ward temperature was maintained at 22°C. All videos were captured at a resolution of 384 x 288 pixels and at 6 frames per second.

Each video was watched by 2 ICU registrars, who then labelled the 23 NAS activities as occurring (True) or not occurring (False) in the video. The registrars were blinded to the patient details, and the infrared video allowed patient privacy to be maintained.

### B. Data Exploration and Preprocessing

The videos had between 676 to 820 frames (mean = 761.59) and were 112 to 137 seconds long. In the 882 videos, 14 out of 23 NAS activities were observed. The remaining 9 NAS activities were not observed in any video. The 'Before' column of Table 1 shows the 14 NAS activities with their number of occurrences observed before preprocessing.

Given that a video may be labelled True for 1 or more NAS activity, this was a multilabel classification problem with 14 classes. To reduce the complexity of the problem, we simplified the problem into a single label classification problem by disregarding labels with less than 50 occurrences and only using videos with exactly 1 occurrence. After preprocessing, we are left with 458 videos from 8 classes (Table 1).

### C. Model Background

We propose the use of the Multiscale Vision Transformer v2 (MViTv2) model for NAS prediction. Proposed by Yanghao et al. in 2022, MViTv2 is a state-of-the-art model that achieved high accuracies on video classification tasks using the benchmark datasets Kinetics-400, Kinetics-600, Kinetics-700 and Something-Something-v2 [20].

MViTv2 is an improved version of a multiscale vision transformer (MViT), whose main idea is to combine Vision Transformers (ViT) with multiscale processing [20, 21]. Firstly, a ViT divides an image into a sequence of patches and applies a transformer model to it [21, 22]. Next, multiscale processing is done by increasing the number of channels while reducing spatial resolution [21]. This is equivalent to the convolutional and pooling layers in a Convolutional Neural Network (CNN) [23]. The idea is that a lower resolution reduces computational power while allowing the model to capture higher-level features [21, 23].

To put our results into perspective, we compared the performance of MViTv2 with 2 other types of video classification models, a 3 Dimensional-CNN (3D-CNN) and a CNN-Recurrent Neural Network (CNN-RNN). Traditionally, video classification has been done using CNN model combinations such as 3D-CNN and CNN-RNN models [24]. This is because typical CNNs are 2-dimensional and can only extract spatial features, which make them unsuitable for processing videos [25]. In a 3D-CNN, a 3-dimensional kernel is used to perform convolutions, capturing both spatial and temporal features [24, 25]. In a CNN-RNN model, a CNN is used to convert each frame into a feature vector [24, 25]. Then, a time series of feature vectors are input into a Recurrent Neural Network (RNN) model which can learn the temporal relationship between the frames [24, 25].

## IV. EXPERIMENTS

In our experiments, we used 2 methods to predict the NAS: 1) indirectly using predicted activity to infer the NAS; and 2) directly predicting the NAS.

### A. Indirectly using predicted activity to infer NAS

For the indirect method, we used a MViTv2 to predict the NAS activity, then inferred the NAS using the Average NAS values from Table 1.

We used Pytorch 2.0's implementation of MViTv2 pretrained on the Kinetics-400 dataset [26]. To customise it for our use, we added a Rectified Linear Unit (ReLU) activation function, then a final dense layer mapping the original 400 possible outputs to 8 possible outputs. The 8 possible outputs represent the prediction of the 8 possible NAS activities.

Since pretraining was done using 16 frames for input, our videos were converted into 16 frames to use the pretrained weights. As our shortest video was 676 frames long, the spacing between the 16 frames was made as large as possible (42 frames) to ensure that these 16 frames are representative of the video. Hence for each video, the central 672 frames were taken and 16 frames with frameskip of 42 were taken from the 672 frames.

These 16 frames serve as input to MViTv2, which will output a number between 0 to 7 representing 1 of the 8 NAS activities.

### B. Directly predicting NAS

For the direct method, we used the MViTv2 to predict the NAS score directly. Similar to the indirect method, we used Pytorch 2.0's implementation of MViTv2 pretrained on the Kinetics-400 dataset [26]. We added a ReLU activation function, then a final dense layer mapping the original 400 possible outputs to 1 possible output. The 1 output represents the NAS.

In the same way as the indirect method, we selected 16 frames from each video as input and the MViTv2 will output the predicted NAS. The loss function used was PyTorch 2.0's MSELoss with reduction set to 'sum'.

### C. Comparison of MViT with R(2+1)D and ResNet50-LSTM

Finally, we compared both methods for 3D-CNN and CNN-RNN model types. We used the R(2+1)D and ResNet50-LSTM models which are generic examples of each type respectively.

In our R(2+1)D model, we used Pytorch 2.0's implementation of R(2+1)D pretrained on the Kinetics-400 dataset [26]. For input, videos were converted into 16 frames in an identical way as above. Using our R(2+1)D, the NAS was predicted both directly and indirectly.

In our ResNet50-LSTM model, ResNet50 was used to 'encode' the video into a feature vector, and a LSTM 'decoded' the feature vector to give the desired output. We used Pytorch 2.0's implementation of ResNet50 with 'IMAGENET1K_V2' weights [26]. For our LSTM, we used the settings input_size = 2048, hidden_size = 256 and num_layers = 2. Like before, each video was converted into 16 frames for input. ResNet50 was used to encode the 16 frames into 16 2048-dimension feature vectors. Then, the LSTM took the feature vectors as input and output the predicted NAS activity or NAS.

During all training phases, we used 5-fold cross validation with 30 epochs in each fold and a batch size of 3. All models were trained on a Titan V GPU using Adam optimizer with learning rate = 0.00003.

To evaluate the performance of our models, we calculated the accuracy (1), the area under the receiver operating characteristic curve (ROC AUC), the macro-averaged F1 score and the mean squared error (MSE) (2) for the indirect method, and the MSE for the direct method. The ROC AUC is the average AUC of the 8 curves given by plotting True Positive Rate (TPR) against False Positive Rate (FPR) for each class. The macro-averaged F1 score is the average F1 score across all classes, where each F1 score is calculated using (3). To infer predicted and true NAS in (2), the Average NAS of the activity is used (Table 1).

$$Accuracy = \frac{No.of\ correctly\ predicted\ videos}{Total\ no.of\ videos} \quad (1)$$

$$MSE = \frac{1}{N} \cdot \sum_{i}^{N}(predicted\ NAS - true\ NAS)^2\ of\ ith\ video \quad (2)$$

$$F1\ score = \frac{2 \cdot true\ positives}{2 \cdot true\ positives + false\ positives + false\ negatives} \quad (3)$$

## IV. RESULTS AND DISCUSSION

Table 2: Accuracy, ROC AUC, macro-averaged F1 score and MSE of NAS prediction using MViTv2, R(2+1)D and ResNet50-LSTM models. Results shown are the average across 5 folds.

**Indirect method**

|  | Accuracy | ROC AUC | F1 score | MSE |
|---|---|---|---|---|
| MViTv2 | 57.21% | 0.865 | 0.570 | 28.16 |
| R(2+1)D | 56.12% | 0.849 | 0.550 | 28.70 |
| ResNet50-LSTM | 51.72% | 0.816 | 0.498 | 34.01 |

**Direct method**[a]

|  | Accuracy | ROC AUC | F1 score | MSE |
|---|---|---|---|---|
| MViTv2 | - | - | - | 18.16 |
| R(2+1)D | - | - | - | 18.45 |
| ResNet50-LSTM | - | - | - | 26.63 |

[a] The direct method does not have an accuracy, ROC AUC or F1 score as the model outputs a numerical score. Only MSE can be calculated which tells us the closeness of the predicted score to the true score.

The MViTv2 achieved the best 5-fold results across all metrics, performing better than the R(2+1)D and ResNet50-LSTM models (Table 2). MViTv2 achieved an accuracy of 57.21%, ROC AUC score of 0.865, F1 score of 0.570 and MSE of 28.16 for indirect prediction and MSE of 18.16 for direct prediction. In comparison, the R(2+1)D was second-best with an accuracy of 56.12%, ROC AUC of 0.849, F1 score of 0.550 and MSE of 28.70 for indirect prediction and MSE of 18.45 for direct prediction. Finally, ResNet50-LSTM came in last with an accuracy of 51.72%, ROC AUC of 0.816, F1 score of 0.498 and MSE of 34.01 for indirect prediction and MSE of 26.63 for direct prediction. This suggests that the MViTv2 is a more effective model for NAS prediction compared to the R(2+1)D and ResNet50-LSTM.

Comparing indirect and direct predictions of NAS, we see that the direct prediction is more precise since its MSE is consistently lower. By bypassing the need for predicting NAS activity, direct prediction is more streamlined and allows the model to predict any NAS value, giving it higher precision. In contrast, the indirect prediction has to decide on the NAS activity in the video, and can only output 8 specific NAS values corresponding to the 8 possible NAS activities. This makes indirect prediction less precise and increases its MSE.

## V. CONCLUSION

In this letter, we demonstrate 3 video classification models for predicting the NAS, propose a novel approach to passively and automatically predict caregiver workload, and ultimately show superior performance of MViTv2. This approach of workload monitoring has the added benefit of anonymizing staff and patients through the use of low-resolution thermal videos.

We also experimented with a novel method of directly predicting the NAS. By not recording specific activity but only deriving a score, no personal information is saved, maintaining privacy while still providing the ability to continuously and anonymously monitor workload.

Other future directions include predicting the NAS for videos with more than 1 activity, and predicting other NAS activities or specific NAS activity subcategories. All these will help to increase the scope and accuracy of NAS prediction.